%% file: main.tex
\documentclass[runningheads]{llncs}

\usepackage{eccv}

\usepackage{eccvabbrv}

\usepackage{graphicx}
\usepackage{booktabs}

\usepackage[accsupp]{axessibility}  %

\usepackage{hyperref}

\usepackage{orcidlink}

\usepackage{tabularx} 
\usepackage{multirow}

\newcommand{\shortname}{LVSPM\xspace}
\newcommand{\methodname}[1]{LVSPM}

\colorlet{cr}{black}

\begin{document}

\title{LVSPM: Long Sequence View Synthesis \\ and Pose Estimation Model} 

\titlerunning{LVSPM}

\author{Xi Chen \inst{1}  \and
Yachi Zhang\inst{2}  \and
Linghao Chen\inst{2}  \and
Minghua Liu\inst{2} \and \\
Hao Su\inst{2} \and
Zexiang Xu\inst{2} \and
Xiaoshuai Zhang\inst{2} }

\authorrunning{X.~Chen et al.}

\institute{UC San Diego \and Sudo AI GmbH
}

\maketitle

\input{sections/0-abstract}
\input{sections/1-intro}

\input{sections/2-related}

\input{sections/3-methods}

\input{sections/4-exps}

\input{sections/5-conclusion}

\bibliographystyle{splncs04}
\bibliography{main}
\end{document}

%% file: sections/0-abstract.tex
\begin{abstract}

We present LVSPM, a generalizable model that jointly estimates camera poses and synthesizes novel views from uncalibrated image collections. Trained with only RGB images and pose supervision, LVSPM avoids dense 3D ground truth and employs test-time training (TTT) layers to scale seamlessly to hundreds of input views.
On RealEstate10k, Co3Dv2, and DL3DV, LVSPM surpasses VGGT in pose estimation across 16–256 views, with especially large margins at strict thresholds. For novel view synthesis under a practical protocol where more views cover larger scenes, LVSPM achieves state-of-the-art pose-free quality---surpassing even pose-dependent models in PSNR---and still maintains high quality as scene scale grows, while baselines collapse. \textcolor{cr}{The code is available at \href{https://burningdust21.github.io/Projects/LVSPM/}{https://burningdust21.github.io/Projects/LVSPM/}.}
\keywords{Novel View Synthesis \and Pose Estimation \and Test-Time Training}
\end{abstract}

%% file: sections/1-intro.tex
\section{Introduction}

Novel view synthesis (NVS) has been a long-standing challenge and a crucial driving force in 3D vision and graphics research. From per-scene optimization-based pipelines \cite{nerf,instantngp,chen2022tensorf,zipnerf,3dgs,mipsplat,2dgs} to feed-forward approaches \cite{pixelnerf,mvsnerf,pixelsplat,mvsplat,hong2024lrm,gslrm,jin2025lvsm}, the community has developed diverse 3D representations and model architectures in recent years in pursuit of increasingly efficient, accurate, and scalable solutions. However, nearly all existing methods assume access to known camera poses, typically estimated through structure-from-motion (SfM) pipelines \cite{schonberger2016structure,snavely2006photo}, which are computationally expensive and often fragile in real-world scenarios.

To avoid the reliance on known poses, several recent attempts have explored pose-free view synthesis with feed-forward models \cite{ye2024nopose,zhang2025flare,jiang2025anysplat}. While promising, these methods are mostly restricted to sparse-view input and cannot scale to long sequences, which are crucial for capturing wide scene coverage and enabling immersive experiences. In parallel, geometry-based models such as DUSt3R and VGGT \cite{wang2024dust3r,wang2025vggt} have shown strong feed-forward performance on pose estimation, but they do not address view synthesis and rely on dense 3D supervision such as point maps that are far more expensive and less scalable to obtain than the image supervision used in classical view synthesis.

\begin{figure*}[t]
    \centering
    \includegraphics[width=\linewidth]{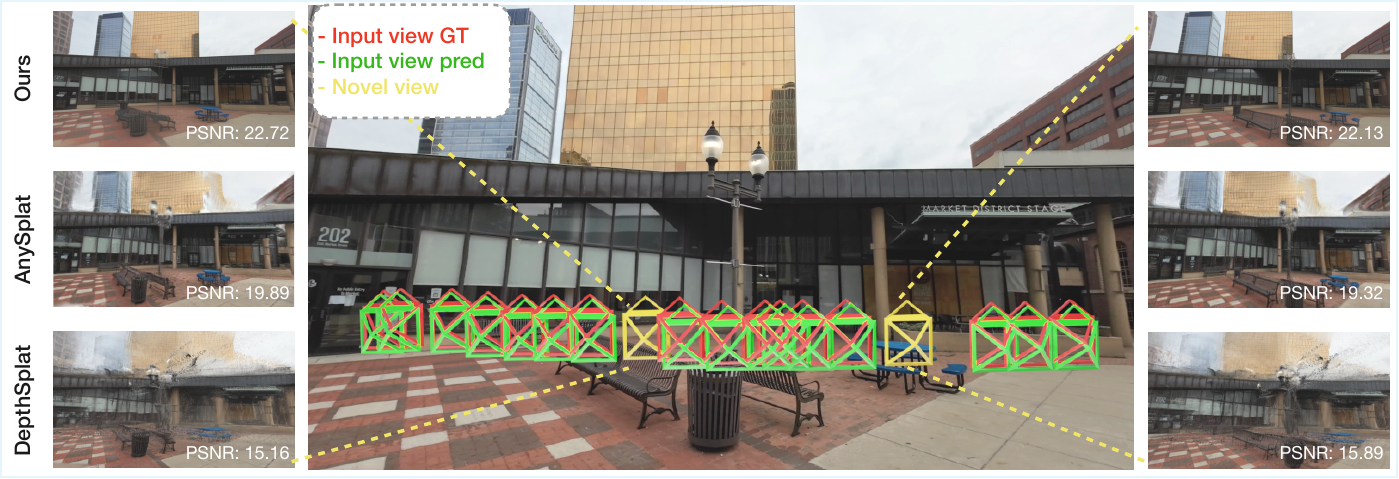}
     \caption{\textbf{ LVSPM} can jointly achieve high-quality novel view synthesis and accurate camera pose estimation from uncalibrated long-sequence (128) multi-view images. A subset of the 128 input views with corresponding prediction and ground-truth poses is shown in the middle, while novel view synthesis comparisons are presented on the side. Our estimated poses closely align with ground truth, while our pose-free synthesis results surpass recent baselines, including the pose-free AnySplat~\cite{jiang2025anysplat} and even the pose-required DepthSplat~\cite{xu2025depthsplat}.}
     
    \label{fig:teaser}
\end{figure*}

This leaves open the challenge of achieving long-sequence pose-free view synthesis without requiring heavy dense 3D labels. %
In this work, we address this challenge by introducing \textbf{a novel feed-forward framework, \shortname{}, that jointly achieves long-sequence novel view synthesis and camera pose estimation from unposed inputs}, using only RGB image and camera pose supervision. As illustrated in Fig.~\ref{fig:teaser}, our approach produces high-fidelity novel renderings and accurate camera poses from hundreds of captured images of a large real scene, achieving practical and scalable long-sequence view synthesis with integrated camera calibration.

To this end, our design philosophy follows the spirit of LVSM~\cite{jin2025lvsm}, aiming to minimize 3D inductive biases by framing the task as sequence-to-sequence token prediction. 
Specifically, given a sequence of unposed input views, our model tokenizes the images and directly predicts tokens corresponding to both input-view camera parameters and novel-view RGB images, without relying on explicit 3D representations or handcrafted geometric modules. 
To scale this to long sequences, we adopt a LaCT backbone~\cite{zhang2025test}---a recent test-time training (TTT) architecture that performs large-chunk TTT updates for efficient and scalable long-sequence modeling. In addition, inspired by VGGT, we incorporate per-view camera tokens into the LaCT architecture, enabling direct pose estimation within the same sequence modeling framework. Together, these components lead to a novel feed-forward system that unifies long-sequence view synthesis and pose estimation within a minimal-bias, scalable framework.

We train our model on multiple large-scale synthetic and real datasets using only RGB images and camera pose supervision, and demonstrate effective scalability to 256 input views at inference. Our contributions are as follows:
\begin{itemize}
    \item We introduce \shortname{}, the first pose-free view synthesis model that scales to 256 input views, enabled by a LaCT backbone with learnable camera tokens that jointly predicts camera poses and novel views in a unified sequence-to-sequence framework.
    \item We demonstrate that novel view synthesis supervision provides sufficient geometric signal for accurate pose estimation, eliminating the need for dense 3D ground truth. Our ablation confirms this: removing NVS supervision reduces the AUC3 from 71.9 to 50.9, while the full model surpasses geometry-based methods trained with dense point-map supervision.
    \item We propose the equi-temporal evaluation protocol that isolates scene-scale effects from sampling density, providing a more realistic benchmark for long-sequence methods.
    \item \shortname{} achieves state-of-the-art results on RE10k, Co3Dv2, and DL3DV for both long sequence pose estimation and view synthesis, surpassing even pose-dependent baselines such as DepthSplat~\cite{xu2025depthsplat}, while scaling gracefully to large scenes where prior methods collapse.
\end{itemize}
These findings highlight the feasibility of addressing fundamental 3D vision problems with cheaper supervision, opening new possibilities for scalable and generalizable 3D perception systems.

%% file: sections/2-related.tex
\section{Related Work}

\textbf{Novel View Synthesis.}
View synthesis has been extensively studied for decades in computer vision and graphics \cite{levoy1996light,gortler1996lumigraph,debevec1996modeling,buehler2001unstructured,zhou2018stereo,nerf,3dgs,hong2024lrm,sajjadi2022srt,jin2025lvsm}.
In recent years, NeRF~\cite{nerf}, 3D Gaussian Splatting(3DGS)~\cite{3dgs}, and many variants of such 3D representations~\cite{instantngp,pointnerf,sun2022direct,chen2022tensorf,fridovich2022plenoxels,2dgs,mipsplat} with differentiable rendering have achieved photo-realistic results through end-to-end optimization, albeit at the cost of significant computational overhead per scene.
To enable faster inference, numerous feed-forward methods have been developed for instant scene reconstruction and rendering, most of which rely on 3D representations and 3D-related architectural designs such as plane-sweep volumes~\cite{mvsnerf,geonerf,zhang2022nerfusion,mvsplat,mvsgaussian} or epipolar priors~\cite{pixelnerf,ibrnet,pixelsplat,gprn}.
Recently, large reconstruction models (LRMs)~\cite{hong2024lrm,li2023instant3d,wang2023pflrm,gslrm,wei2024meshlrm} have begun to reduce such architectural-level 3D biases, leveraging pure transformer architectures, while LVSM~\cite{jin2025lvsm} further eliminates representation-level bias and achieves state-of-the-art view synthesis quality.
Our method inherits this minimal-bias philosophy but extends it to the joint problem of view synthesis and pose estimation, whereas most existing approaches still assume known camera poses.

On the other hand, most prior feed-forward methods remain restricted to sparse input views. Recent works such as Long-LRM~\cite{ziwen2024long} and LaCT~\cite{zhang2025test} have begun to explore long-sequence inputs with tens or even hundreds of images using architectures like Mamba~\cite{gu2023mamba,dao2024mamba2} and TTT~\cite{sun2024learning}; however, these approaches still assume known camera poses, which limits their practicality. Our approach builds upon LaCT and extends it to the pose-free setting, enabling joint long-sequence view synthesis and pose estimation.

\textbf{Camera Pose Estimation.}
Camera pose estimation has traditionally relied on Structure-from-Motion (SfM) pipelines~\cite{schonberger2016structure,snavely2006photo}, which remain robust for large-scale reconstructions but suffer from heavy computational cost and failure modes in textureless regions or repetitive structures. These issues persist even with the advent of learning-based feature extractors~\cite{detone2018superpoint,dusmanu2019d2,revaud2019r2d2} and matchers~\cite{sarlin2020superglue,sarlin2019hloc,liu2021learnable}. Recently, many learning-based approaches have attempted to directly regress camera poses~\cite{lin2023relposepp,rockwell20228,cai2021extreme}, though the current state of the art comes from geometry-driven transformer-based models \cite{wang2024dust3r,wang2025vggt, lin2025depth}. In particular, DUSt3R~\cite{wang2024dust3r} formulates pairwise 3D reconstruction as point map regression, enabling pose-free 3D reconstruction and subsequent camera estimation via PnP. Numerous follow-up works have extended DUSt3R, improving its inference quality and scalability to longer sequences~\cite{yang2025fast3r,yuan2025test3r,tang2025mv,wang2025continuous,leroy2024grounding, chen2025ttt3r}. More recently, \textcolor{cr}{VGGT~\cite{wang2025vggt} represents a major step forward, introducing a feed-forward multi-task transformer that jointly predicts cameras, depth, and point maps, improving performance significantly across several geometry tasks, including pose estimation. DepthAnything3~\cite{lin2025depth} extends the feed-forward visual geometry paradigm of VGGT with a simpler plain-transformer architecture and a unified depth-ray prediction target, together with a teacher-student training strategy for dense geometry supervision, achieving state-of-the-art performance.} However, these approaches primarily focus on geometric reconstruction, depend heavily on dense 3D labels, and do not directly address the problem of view synthesis. Our method takes inspiration from VGGT’s camera estimation mechanism but integrates it into a view synthesis framework, achieving comparable or superior pose estimation results while eliminating the need for dense geometric ground truth.

\textbf{Pose-Free View Synthesis.}
Many recent works have sought to remove the requirement of known camera poses in view synthesis to improve practicality. Early attempts integrated pose estimation into optimization-based frameworks such as NeRF, jointly estimating poses and scene representations during training \cite{lin2021barf,bian2023nope,chen2023local,truong2023sparf,xia2022sinerf}.
\textcolor{cr}{Among them, RayZer~\cite{jiang2025rayzer} studies self-supervised novel view synthesis from unposed images by using self-predicted camera poses for rendering. However, these cameras are primarily designed to support view synthesis rather than being directly formulated as world-aligned pose outputs. In contrast, our method jointly estimates world-aligned cameras and synthesizes novel views over long image sequences, scaling to hundreds of input views.}
Recently, feed-forward approaches have gained traction. While some methods adopt a two-stage pipeline that first estimates cameras and then performs view synthesis \cite{jiang2022forge,jiang2025rayzer,zhang2025flare}, others aim to predict poses and novel views jointly. Early pose-free approaches typically achieve only sparse-view, object-level reconstruction and rendering~\cite{wang2023pflrm,jiang2023leap,xu2024sparp}.
Building on the success of DUSt3R and 3D Gaussian Splatting, subsequent works have extended this idea to scene-level pose-free view synthesis by directly reconstructing Gaussian point clouds from unposed inputs and recovering poses via PnP~\cite{ye2024nopose,smart2024splatt3r,xu2024freesplatter,fan2024instantsplat,huang2025no}. However, these approaches generally rely on dense geometric supervision and remain limited to only a handful of input views (often fewer than ten).
More recently, AnySplat~\cite{jiang2025anysplat} extends VGGT to the view synthesis task, supporting several tens of input views, but it depends on VGGT pretraining and still requires dense 3D labels for supervision. In contrast, our method outperforms AnySplat under its setting and scales effectively to hundreds of input images. Notably, our model is trained from scratch using only RGB image and camera pose supervision, without requiring additional dense 3D geometric labels.

%% file: sections/3-methods.tex
\section{Method}
Our work, \methodname{}, introduces a feed-forward model designed to jointly estimate camera poses and synthesize novel views from a sequence of unposed images. The model architecture minimizes explicit 3D inductive biases, framing the problem as a sequence-to-sequence prediction task. We leverage a Large-Chunk Test-Time Training (LaCT) backbone~\cite{zhang2025test} to efficiently process long input sequences, coupled with learnable camera tokens for direct camera pose and intrinsic parameter regression. This section details the model's input representation, architecture, and the supervision strategy used for training.

\subsection{Problem Formulation and Overview}

Given a set of $N$ uncalibrated input images $\mathcal{I} = \{I_i\}_{i=1}^N$, where $I_i \in \mathbb{R}^{H \times W \times 3}$, our goal is to:
\begin{itemize}
    \item Estimate camera poses $\{E_i=(t_i, r_i)\}_{i=1}^N$, as well as intrinsic parameters $\{K_i=(f^x_i, f^y_i)\}_{i=1}^N$, where $t_i \in \mathbb{R}^3$ is the 3-dim translation, $r_i \in \mathbb{R}^4$ is the 4-dim quaternion, and $\hat{f}^x_i, \hat{f}^y_i$ are scalars denoting the horizontal and vertical camera fields of view in radians, respectively;
    \item Synthesize novel view $I^t$ from target novel camera configurations $(E^t, K^t)$.
\end{itemize}
This joint formulation eliminates the dependency on pre-computed structure-from-motion (SfM) pipelines while enabling end-to-end learning of geometric and appearance modeling.

\begin{figure*}[t]
    \centering
    \includegraphics[width=\linewidth]{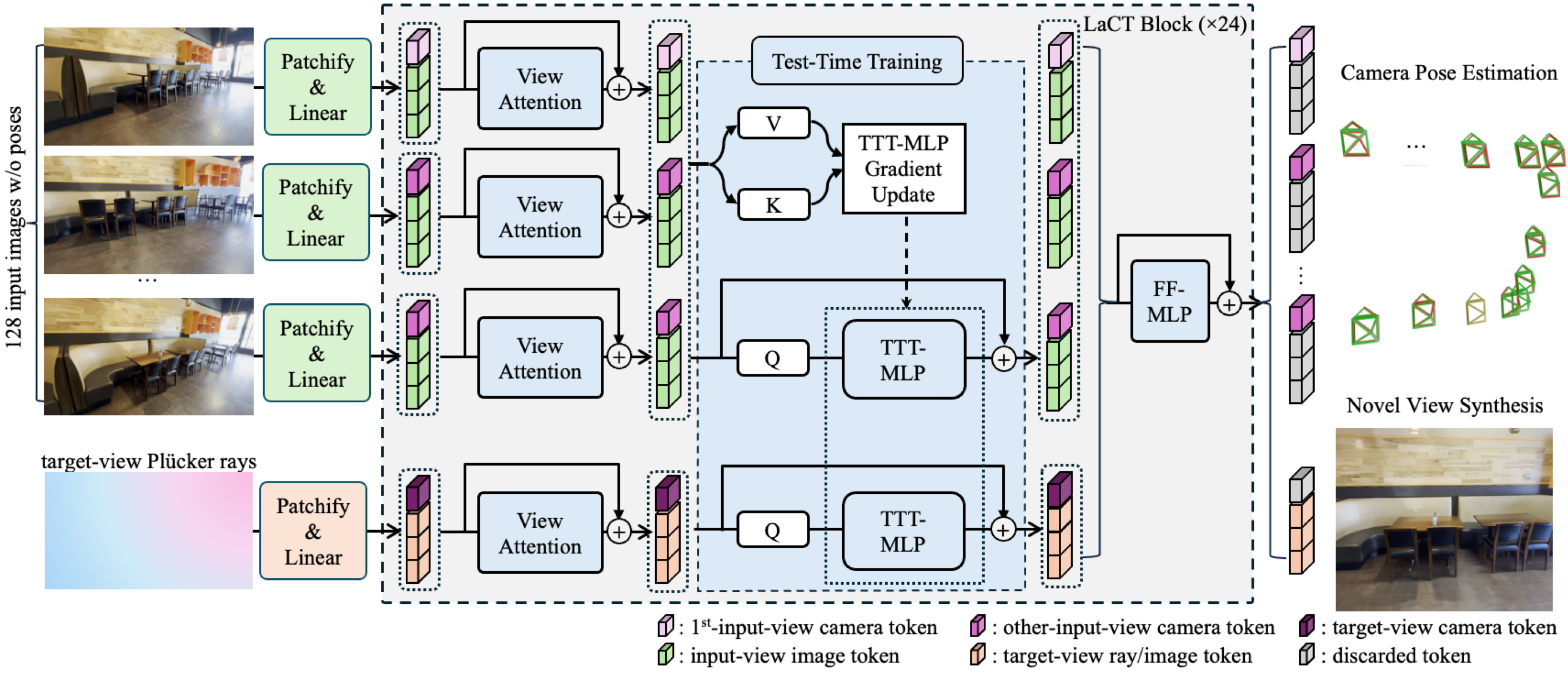}
     \caption{\textbf{Pipeline.} Our approach takes uncalibrated long-sequence input views, where each image is tokenized into patch tokens and augmented with a learnable camera token, while target views are represented by Plücker-ray tokens. All tokens are processed through a stack of LaCT blocks that combine per-view self-attention, an MLP-based TTT layer, and a feed-forward MLP layer. From the final tokens, our lightweight decoders produce novel-view RGB images and camera parameters, enabling accurate pose estimation together with high-quality long-sequence view synthesis.}
    \label{fig:pipeline}
\end{figure*}

As shown in Fig.~\ref{fig:pipeline}, \methodname{} consists of $L$ stacked LaCT blocks, each containing three main components: (1) a large-chunk test-time training (TTT) layer for long-range multi-view reasoning, (2) a window attention layer for local spatial dependencies, and (3) a feed-forward MLP network for channel mixing. The architecture processes variable-length sequences up to 1M tokens through an efficient test-time adaptation mechanism. 

\subsection{Model details}

\textbf{Input representation and tokenization.} Unlike prior pose-required view synthesis models~\cite{gslrm,jin2025lvsm} that typically condition both input and target views on Plücker rays~\cite{plucker1865xvii}, our pose-free approach ignores input-view Plücker encoding and instead introduces additional learnable camera tokens to extract camera information. Specifically, each input image $I_i$ is divided into non-overlapping patches of $p \times p$ pixels, noted as $\{I_{ij}\in\mathbb{R}^{p\times p \times 3}| j =1,\ldots,HW/p^2\}$. These patches are then directly projected to the model dimension $d$ with a linear layer: $x_{ij} = \operatorname{Linear}_\textit{input}(I_{ij})$. Inspired by VGGT~\cite{wang2025vggt}, we append each input image with a learnable camera token $c_i$ for pose prediction and intra-view context exchange. The first view is used as the canonical reference frame, assuming a zero translation and an identity rotation, and has a special initial learnable camera token $c_r$ to represent the reference frame. All other input views share another learnable camera token $c_x$. \textcolor{cr}{To ensure a consistent coordinate system across scenes, the translation scale is normalized by setting the distance between the first and furthest points to unit length. This scene normalization, derived from input views, is then applied to target views, aligning both sets of views in a consistent coordinate space.}

The input-view tokens $X_i$ for input view $i$ are represented as:
\begin{align}
X_i = \{c_i, x_{i1}, x_{i2}, \ldots, x_{iM}\}, \
c_1 = c_r, \quad c_i = c_x \ \text{for} \ i=2,\ldots,N.
\end{align}
where $M=HW/p^2$. For simplicity, we denote by $X_{ij}$ an arbitrary token from input view $i$, where $X_{i0} = c_i$ corresponds to the camera token, and $X_{ij} = x_{ij}$ for $j > 0$ corresponds to the image tokens. 

On the other hand, we still adopt Plücker rays to condition target views, providing the 3D camera context required for novel-view synthesis. More specifically, we compute the target frame Plücker rays from its camera parameters $(E^t,K^t)$. For the $k$-th target view, its Plücker ray patches are noted as $P^t_{k} = \{P^t_{kj} | j = 1,\ldots,HW/p^2\}$. These Plücker ray patches are transformed into tokens using another linear layer, similar to how we encode the image patches of the input views: $x^t_{kj} = \operatorname{Linear}_\textit{target}(P^t_{kj})$.
To be consistent with input view encoding, these target tokens are concatenated with another learnable token $c_t$ before feeding into the network. The target view tokens $T_k$ for target view $k$ is represented as:
\begin{align}
    T_k = \{c_k, x^t_{k1}, x^t_{k2}, \ldots, x^t_{kM}\}, \quad c_k=c_t
\end{align}
Similar to input-view token $X_{ij}$, we denote by $T_{kj}$ an arbitrary token from target view $k$.

\textbf{LaCT blocks.}
We adopt the recent Large-Chunk Test-Time Training (LaCT) architecture~\cite{zhang2025test} to process both input and target view tokens for joint view synthesis and pose estimation, effectively handling unordered, unposed multi-view inputs while remaining computationally efficient.
In particular, our model consists of $L=24$ LaCT blocks with alternating (windowed) self-attention, test-time training (TTT), and Feed-forward MLP layers; each TTT layer is equipped with a SwiGLU-MLP~\cite{shazeer2020glu}, noted as $\textbf{TTT-MLP}_l$, for test-time training. 

Specifically, for the $l$-th block, the $HW/p^2 + 1$ tokens of each view are first fed into a per-view windowed self-attention layer: %
\begin{align}
    \hat{X}^l_i &= \operatorname{Attn}_{l}(X^{l-1}_i)+X^{l-1}_i, \\ 
    \hat{T}^l_k &= \operatorname{Attn}_{l}(T^{l-1}_k)+T^{l-1}_k.
\end{align}
Then we process multi-view tokens with the TTT layer, specifically:
\begin{align}
    q_{l,ij} = \operatorname{Q}_l(\hat{X}^l_{ij}),& \quad q^t_{l,kj} = \operatorname{Q}_l(\hat{T}^l_{kj}), \\ 
    k_{l,ij} &= \operatorname{K}_l(\hat{X}^l_{ij}), \\ 
    v_{l,ij} &= \operatorname{V}_l(\hat{X}^l_{ij}),
\end{align}
where $Q_l$, $K_l$, and $V_l$ are learnable parameters to project the original tokens into Q, K, V parameters. The weight of $\operatorname{TTT-MLP}_l$ is then updated with the gradient $G_l = \nabla_{\operatorname{TTT-MLP}_l} (1 -  \operatorname{TTT-MLP}_l(k_l) \cdot v_l^T)$ and a learnable learning weight $\eta$. The updated $\operatorname{TTT-MLP}_l^{\text{updated}}$ is used to process the final outputs of this TTT layer for token update:
\begin{align}
    \Tilde{X}^l_{ij} &= \operatorname{TTT-MLP}_l^\text{updated}(q_{l,ij}) + \hat{X}^l_{ij}, \\
    \Tilde{T}^l_{kj} &= \operatorname{TTT-MLP}_l^\text{updated}(q^t_{l,kj}) + \hat{T}^l_{kj}
\end{align}
Note that only input view tokens are sent through $\operatorname{K}_l, \operatorname{V}_l$ and generate gradient for TTT MLP updates, as done in~\cite{zhang2025test}. This way, the target view tokens don’t need to interact with one another, enabling efficient synthesis of each novel view independently.
The token outputs from the $l$-th LaCT block come from a final feed-forward MLP, denoted as $\operatorname{FF-MLP}_l$:
\begin{align}
    X^l_{ij} &= \operatorname{FF-MLP}_l(\Tilde{X}^l_{ij}) + \Tilde{X}^l_{ij}, \\
    T^l_{kj} &= \operatorname{FF-MLP}_l(\Tilde{T}^l_{kj}) + \Tilde{T}^l_{kj}
\end{align}

\textbf{Design insights.} A key consequence of the asymmetric K/V routing---where only input tokens update TTT weights while target tokens are read-only---is that each novel view can be synthesized independently without interacting with other target views, enabling efficient parallel rendering at 58.8 FPS regardless of the number of targets. The same mechanism naturally supports the joint pose+NVS formulation: the TTT hidden state serves as a compressed scene representation that is queried by both camera tokens (for pose) and target ray tokens (for synthesis). As confirmed by our ablation in Sec.~\ref{sec:ablation}, this joint formulation creates a virtuous cycle where NVS supervision provides rich geometric signal that dramatically improves pose accuracy, eliminating the need for dense 3D supervision used by prior methods~\cite{wang2024dust3r,wang2025vggt}.

\textbf{Prediction Heads.}
After the 24 LaCT blocks, target view tokens are decoded with a two-layer MLP, denoted as  $\operatorname{MLP}_\text{rgb}$, and rearranged to form the final novel-view RGB predictions. 
On the other hand, in contrast to the heavy camera head used by VGGT~\cite{wang2025vggt}, we adopt simple light weight MLPs for camera parameter decoding. Specifically, the camera pose of each view is decoded from the final camera token with a simple two-layer $\operatorname{MLP}_\text{pose}$, where the output is 9-dim: a 4-dim quaternion, a 3-dim translation, and 2-dim $\hat{f}^x$ and $\hat{f}^y$ fields of view concatenated. We show that such lightweight camera heads are already sufficient to produce accurate camera parameters, even surpassing VGGT in estimation accuracy.

Note that VGGT is a transformer-based architecture; to achieve camera estimation, VGGT introduces frame-attention modules coupled with full attention to process the camera token jointly with per-view image tokens. In contrast, our design leverages LaCT blocks, which natively incorporate local windowed attention, eliminating the need for extra modules. As a result, our framework achieves better camera estimation while using fewer model parameters. %

\vspace{-0.5em}
\subsection{Training Objective}
\vspace{-0.5em}

The model is trained with photometric, camera pose, and intrinsic losses:
\begin{align}
\mathcal{L}_{\text{rgb}} &= ||I_\text{pred} - I_\text{gt}||_2^2 + \lambda_{\text{LPIPS}} \cdot \operatorname{LPIPS}(I_\text{pred}, I_{\text{gt}}), \\
\mathcal{L}_{\text{pose}} &= ||t_\text{pred} - t_{\text{gt}}||_1 + ||r_\text{pred} - r_{\text{gt}}||_1, \\
\mathcal{L}_{\text{int}} &= ||\hat{f}^x_\text{pred} - \hat{f}^x_{\text{gt}}||_1 + ||\hat{f}^y_\text{pred} - \hat{f}^y_{\text{gt}}||_1,
\end{align}
where $I_\text{pred}$, $t_\text{pred}$, $r_\text{pred}$, $\hat{f}^x_\text{pred}$, and $\hat{f}^y_\text{pred}$ are RGB image, camera translation, rotation quaternion, and x, y fields of view predictions respectively, and the subscript $_\text{gt}$ denotes the ground-truth values for the corresponding quantities.

The total training loss is: $$\mathcal{L}_{\text{total}} = \lambda_{\text{rgb}} \mathcal{L}_{\text{rgb}} + \lambda_{\text{pose}} \mathcal{L}_{\text{pose}} + \lambda_{\text{int}} \mathcal{L}_{\text{int}}.$$
where $\lambda_{\text{rgb}}$, $\lambda_{\text{pose}}$, and $\lambda_{\text{int}}$ are the weights for the RGB, pose, and intrinsic losses, respectively.

%% file: sections/4-exps.tex
\section{Experiments}

We evaluate LVSPM on multiple challenging benchmarks to demonstrate its effectiveness at joint camera pose estimation and view synthesis from unposed input views. For pose estimation, we evaluate on RealEstate10k (RE10k)~\cite{zhou2018stereo}, Co3Dv2~\cite{reizenstein21co3d}, and DL3DV-10K~\cite{ling2024dl3dv}. For novel view synthesis, we evaluate on  DL3DV-10K~\cite{ling2024dl3dv}, and zero-shot on Tanks and Temples~\cite{Knapitsch2017} and MipNeRF360~\cite{barron2022mipnerf360}, comparing against both pose-free and pose-dependent baselines. We evaluate our method across varying numbers of input views- from sparse (16 views) to large-scale real-world sequences with 256 views.
Our experiments validate three key claims: (1) LVSPM achieves state-of-the-art pose-free view synthesis quality, (2) our pose estimation matches or exceeds specialized geometry-based methods despite using no dense 3D supervision, and (3) the model scales gracefully from small- to large-scale scenes.

\subsection{Experiment Settings}

\textbf{Model Details.} 
Our model uses 24 layers of alternating view-attention and test-time-training layers. Model dimension $d=768$. We use 12 heads for self-attention layers. We follow~\cite{zhang2025test} to apply RoPE embedding~\cite{su2024roformer} on $\operatorname{Q}$ and $\operatorname{K}$ inputs. The model contains 312M parameters.

\textbf{Training Details.}
 We implement LVSPM with PyTorch and train on 64 NVIDIA H100 GPUs. We train our model on a mix of large-scale synthetic and real-world datasets, including Aria Synthetic Environments (ASE)~\cite{Pan_2023_ICCV}, DL3DV-10K~\cite{ling2024dl3dv}, ScanNet++~\cite{yeshwanthliu2023scannetpp}, Hypersim~\cite{roberts2021hypersim}, and CO3Dv2~\cite{reizenstein21co3d}. Our model is not trained on the evaluation split of these datasets. The model undergoes a three-stage training process. First, we pre-train on the synthetic ASE dataset for 90k iterations with 32 input and target views at $128 \times 128$ resolution. Second, we mix ASE with other training datasets (DL3DV-10K, ScanNet++, Hypersim, Co3Dv2), and train for another 60k iterations. Finally, we progressively increase input resolution to $512 \times 288$, and increase scale to 128 input and 64 target views.
 Although training uses at most 128 views, the TTT mechanism naturally generalizes to longer sequences at inference: its hidden state is updated once with all input tokens, so processing 256 views simply extends the update sequence without architectural changes. We verify this empirically---our 256-view results remain strong without any long-sequence finetuning.
 More details can be found in the Supplementary material.

\input{tables/pose-main}

\textbf{Baselines.}
We compare against recent pose estimation methods, namely Fast3R~\cite{yang2025fast3r}, Cut3R~\cite{cut3r}, Flare~\cite{zhang2025flare}, VGGT~\cite{wang2025vggt}, \textcolor{cr}{DepthAnything3 (DA3)~\cite{lin2025depth}, and TTT3R~\cite{chen2025ttt3r}}. Details for resolutions are in the supplementary. For view synthesis, we compare with both pose-free method AnySplat~\cite{jiang2025anysplat} and pose-dependent method DepthSplat~\cite{xu2024depthsplat}. Following DepthSplat~\cite{xu2024depthsplat}, we evaluate LVSPM and baselines at a resolution of 448 × 256. \textcolor{cr}{We use ground truth camera poses of the target views for LVSPM and DepthSplat, while Anysplat uses its predicted camera poses to follow their evaluation protocol. We include a comparison using the LVSPM predicted target camera pose in the supplementary material for a more comprehensive evaluation.}  
For fair comparison, we use official implementations where available and use the same evaluation datasets for all models.

\textbf{Evaluation Metrics.}
For pose estimation, we follow VGGT~\cite{wang2025vggt} to report Area Under Curve (AUC) metrics at thresholds of 3$^\circ$ (AUC3), 5$^\circ$ (AUC5), and 30$^\circ$ (AUC30). The pose AUC evaluates both rotation and translation accuracy jointly---a pose is considered correct at a given threshold only if both relative rotation error (RRE) and relative translation error (RTE) are below the threshold. Higher AUC values indicate more accurate pose estimation. AUC3 is the strictest metric, requiring sub-3-degree precision for both rotation and translation. For view synthesis, we measure PSNR, SSIM~\cite{Wang2004ssim}, and LPIPS~\cite{zhang2018lpips} on held-out test views, reporting average metrics across scenes. We also report network runtime across 16-256 views.

\subsection{Camera Pose Estimation}
\input{tables/dl3dv-pose}

\textbf{Evaluation Protocol.} We adopt an equi-temporal evaluation protocol that reflects practical capture scenarios: $N$ input views are sampled from the first $N/N_{\max}$ portion of the video sequence, so that more views correspond to larger scene coverage rather than denser sampling of the same area. Within each equi-temporal window, input views are randomly sampled for pose estimation following~\cite{wang2025vggt} and evenly sampled for NVS. This makes the task progressively harder with more views, as the model must handle a larger, more diverse scene.
Standard uniform sampling over the full sequence conflates two variables---scene coverage and sampling density---making it unclear whether improvements come from seeing more of the scene or from denser observations of the same area. Our equi-temporal protocol isolates the effect of scene scale by holding sampling density approximately constant, providing a cleaner evaluation of a model's ability to handle diverse, large-scale environments. We also evaluate under standard sparse-view protocols following prior work~\cite{xu2024depthsplat,ye2024nopose}; these results are provided in the supplementary material and confirm our method's competitiveness across evaluation settings.
For RE10k and Co3Dv2, $N_{\max}=128$ with $N\in\{16,32,64,128\}$. For the large-scale DL3DV dataset, $N_{\max}=256$ with $N\in\{16,32,64,128,256\}$.

Our pose estimation results in Tab.~\ref{tab:pose} and Tab.~\ref{tab:dl3dvpose} demonstrate LVSPM's strong performance across all datasets and view counts, with especially large advantages at strict accuracy thresholds (AUC3 and AUC5).

On RealEstate10k (RE10k), LVSPM dominates nearly all metrics. At 16 views, we achieve 40.0 AUC3 and 53.8 AUC5, compared to VGGT's 14.0 AUC3 and 25.3 AUC5---nearly 3$\times$ and 2$\times$ the precision. At 128 views (full sequence), the gap grows further: 61.3 vs.\ 20.6 AUC3 and 72.5 vs.\ 34.1 AUC5. \textcolor{cr}{Compared with DA3, LVSPM achieves higher AUC${30}$ and AUC$5$ across all view counts and remains competitive on AUC$3$, winning 10 out of 12 metrics, despite using only RGB images and camera poses rather than large-scale dense geometric supervision.} Flare emerges as the third-best method at longer sequences (35.9 AUC3 at 128 views) but remains far behind ours.

On Co3Dv2, LVSPM outperforms VGGT at all metrics. At 16 views, we achieve 39.2 AUC3 vs.\ VGGT's 33.9 and 50.1 AUC5 vs.\ 45.2. At 128 views, the margins remain consistent: 68.8 vs.\ 62.1 AUC3 and 76.6 vs.\ 71.4 AUC5. \textcolor{cr}{DA3 sets a strong pose-estimation reference with dense geometry supervision, especially in a sparse-view setting. LVSPM remains competitive under this comparison and closes the gap as context grows, surpassing DA3 on AUC$3$ at 64/128 views and AUC$_5$ at 128 views while using no dense geometry labels and jointly supporting NVS. This trend supports our design goal of scalable long-sequence aggregation rather than optimizing only short-context pose estimation.}

On the challenging DL3DV dataset featuring large-scale scenes, LVSPM demonstrates both high accuracy and strong scalability. 
\textcolor{cr}{At 16 views, we achieve 88.4 AUC3 vs.\ VGGT's 74.7 (+13.7 absolute) and DA3's 84.0, and 91.6 AUC5 vs.\ 81.9 and 88.7. At 256 views (full sequence), we maintain 90.1 AUC3 vs.\ 83.6 and 83.8,} while other baselines collapse: Fast3R drops to 8.0 AUC3 and Cut3R to 19.2 AUC3. These results validate that joint NVS supervision and the TTT architecture enable both robust long-sequence scaling and fine-grained pose accuracy without dense geometric supervision.
\input{tables/dl3dv-nvs}

\begin{figure*}[h!]
    \centering
    \includegraphics[width=0.90\linewidth]{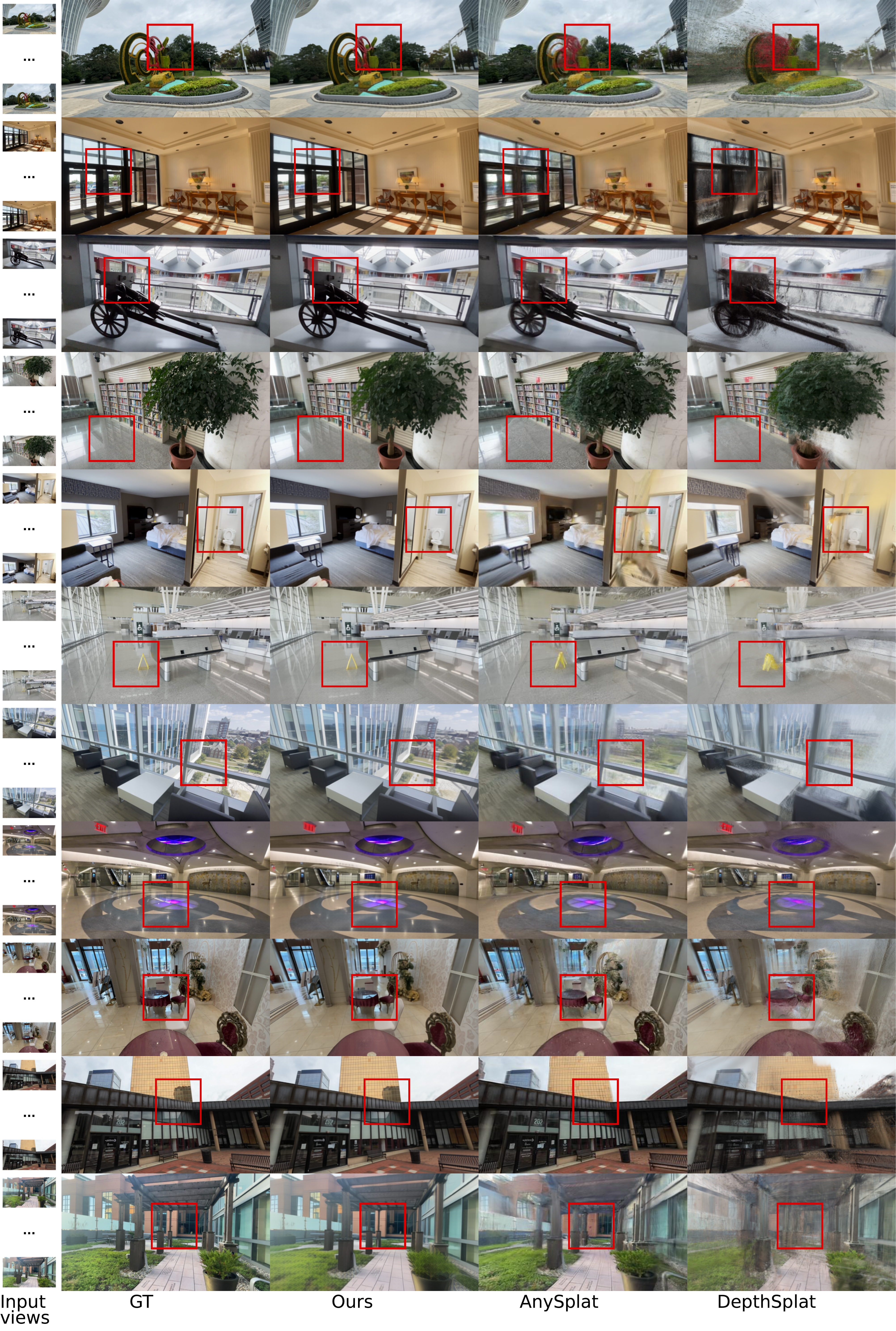}
     \caption{\textbf{Novel-view rendering comparison on the DL3DV dataset with long-sequence input.} We visualize results for 64 inputs using our method, AnySplat and DepthSplat. Unlike DepthSplat, which requires known poses, our method and AnySplat are pose-free. }
    \label{fig:main_exp_dl3dv}
\end{figure*}

\subsection{View Synthesis}

Table~\ref{tab:viewsyn} shows that \methodname{} achieves state-of-the-art pose-free view synthesis on DL3DV-10K. At 16 views (the smallest scene coverage), LVSPM achieves 25.77 PSNR, surpassing even the pose-dependent DepthSplat (24.80) and far exceeding the pose-free AnySplat (23.15). This demonstrates strong synthesis quality in the relatively sparse regime without requiring any pose input.

Under our equi-temporal protocol, increasing the number of input views corresponds to larger scene coverage and hence harder tasks. All methods naturally degrade, but LVSPM shows the most graceful scaling: our PSNR decreases from 25.77 to 22.09 ($-$3.68\,dB) across a 16$\times$ increase in scene scale, while AnySplat drops from 23.15 to 18.55 ($-$4.60\,dB) and DepthSplat collapses from 24.80 to 18.39 at just 64 views ($-$6.41\,dB) before failing due to its model constraints. Critically, the gap between LVSPM and baselines \emph{widens} with scale: +2.62\,dB over AnySplat at 16 views grows to +3.54\,dB at 256 views. LPIPS shows similar robustness (ours 0.152$\rightarrow$0.250 vs.\ AnySplat 0.158$\rightarrow$0.339). Qualitative results in Fig.~\ref{fig:main_exp_dl3dv} confirm that our method renders sharper reflections and avoids the layered-surface artifacts of AnySplat~\cite{jiang2025anysplat}.
\textbf{Zero-shot Evaluation.} We zero-shot generalize to the challenging Tanks and Temples (TnT)~\cite{Knapitsch2017Tanks} dataset and the indoor MipNeRF360 dataset~\cite{barron2022mipnerf360} under both dense (64--128 views) and sparse (3--6 views) input settings. Full quantitative tables and qualitative comparisons are provided in the supplementary material. \methodname{} consistently outperforms baselines across all settings and view counts.

\input{ tables/re10k}

\textbf{Sparse view Evaluation.} We also evaluate \methodname{} under sparse view setting on unseen datasets RE10K~\cite{zhou2018stereo}, MipNeRF360~\cite{barron2022mipnerf360} and LLFF~\cite{mildenhall2019llff}, and compare with both posed (e.g. MVSplat~\cite{mvsplat}) and unposed (e.g. NoPoSPlat~\cite{ye2024nopose}) baselines. We finetune \methodname{} with sparse input views and zero-shot generalize to these datasets. \textcolor{cr}{For RE10K, the results are shown in Table~\ref{tab:re10k}. \methodname{} surpasses AnySplat under the zero-shot setting. When fine-tuned on RE10K (Ours - F.T.), it also outperforms other baselines trained exclusively on this dataset. Full MipNeRF360/LLFF sparse-view results and finetuning details are in the supplementary.
}

\subsection{Ablation Study}
\label{sec:ablation}

We conduct ablations on DL3DV and ASE using 32 input and target views. First, we evaluate the effect of large-scale synthetic pretraining by directly training on real data (``w/o Syn.''). As shown in Table~\ref{tab:ablation}, synthetic pretraining consistently improves both pose estimation and novel-view rendering on DL3DV.

We then validate the model design on the ASE synthetic dataset, due to computational constraints. We first validate the effect of NVS supervision on pose estimation (``w/o RGB''). As shown in Table~\ref{tab:ablation}, removing NVS supervision causes AUC3 to drop from 71.9 to 50.9---a nearly 30\% degradation. This dramatic gap demonstrates that NVS supervision provides rich geometric signal that effectively replaces the dense 3D supervision (point maps, depth) used by methods like VGGT~\cite{wang2025vggt} and DUSt3R~\cite{wang2024dust3r}.

We also ablate the TTT chunk size and the number of TTT layers. Our full model uses a chunk size of 8224 tokens (aggregated from all input images and cameras) and performs one MLP update per TTT layer. In the chunk-size ablation ('Half Chunk size'), we halve the chunk size to 4112 and apply two updates per layer; as shown in Table~\ref{tab:ablation}, this causes a slight drop in NVS and noticeable degradation in pose estimation. We hypothesize that repeated updates on smaller chunks promote catastrophic forgetting, causing TTT layers to overfit local geometry and lose the global context needed for consistent long-sequence alignment. In contrast, a single-pass update over a larger context window is more stable for joint pose–NVS optimization and is also more efficient. We further ablate the number of layers ('12 TTT layers') by reducing from 24 to 12 layers, which causes worse performance in Table~\ref{tab:ablation}, indicating that a smaller TTT capacity is insufficient. Our full models give the best overall performance.

 \subsection{Inference Time comparison} 

 \begin{figure}[t]
\centering
\begin{minipage}[t]{0.58\columnwidth}
\vspace{0pt}
\centering
\resizebox{0.9\linewidth}{!}{%
\begin{tabular}{c|c|ccc|ccc}
\toprule
 &  & \multicolumn{3}{c|}{Pose} & \multicolumn{3}{c}{NVS} \\
Dataset & Method & AUC30 & AUC5 & AUC3 & PSNR & SSIM & LPIPS \\
\midrule
\multirow{2}[2]{*}{DL3DV}
& w/o Syn & \underline{92.7} & \underline{84.3} & \underline{79.9} & \underline{20.08} & \underline{0.589} & \underline{0.267} \\
& Ours    & \textbf{94.5} & \textbf{85.6} & \textbf{80.2} & \textbf{21.12} & \textbf{0.638} & \textbf{0.222} \\
\midrule
\multirow{4}[2]{*}{ASE}
& w/o RGB         & {87.9} & {62.6} & {50.9} & -- & -- & -- \\
& Half Chunk size & 88.3 & \underline{74.3} & \underline{67.9} & \underline{24.98} & \textbf{0.702} & \underline{0.263} \\
& 12 TTT layers   & \underline{90.1} & {69.2} & {59.3} & 24.18 & 0.673 & 0.288 \\
& Ours Full       & \textbf{93.7} & \textbf{79.5} & \textbf{71.9} & \textbf{25.01} & \underline{0.701} & \textbf{0.257} \\
\bottomrule
\end{tabular}%
} %
\captionof{table}{Ablation study on DL3DV-10K and ASE datasets.}
\label{tab:ablation}
\end{minipage}%
\hspace{0.02\columnwidth}%
\begin{minipage}[t]{0.40\columnwidth}
\vspace{0pt}
\centering
\includegraphics[width=\linewidth]{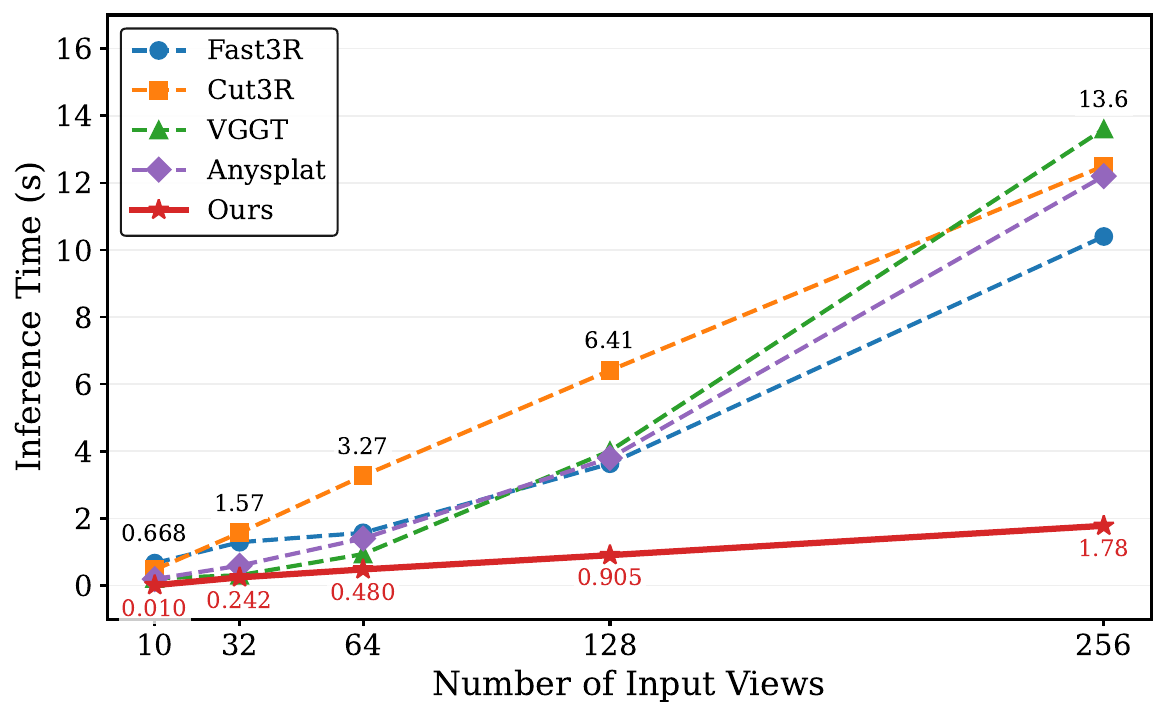}
\captionof{figure}{Inference time comparison.}
\label{fig:time}
\end{minipage}
\end{figure}
\vspace{-0.5em}

 We report inference time for LVSPM and all baselines across 10–256 input views. We use a resolution of 512×288 for LVSPM, Cut3R, Fast3R, and AnySplat, and 518×280 for VGGT due to its 14 patch size. For each method, we measure the time required to predict poses for all input views, using an H100 GPU. As shown in Fig.~\ref{fig:time}, LVSPM scales linearly with the number of input views and processes 256 images in under 2 seconds, whereas baselines require over 10 seconds. Additionally, our method renders at \textbf{58.8} FPS, regardless of the number of input views.

%% file: tables/pose-main.tex
\begin{table*}[t]
  \centering
  \caption{
  Pose estimation results (AUC$\uparrow$) on RE10K and Co3Dv2 with different numbers of input views.
  \textbf{Best} and \underline{second-best} results are marked in bold and underlined. Despite no dense geometry supervision, LVSPM is best or second-best in all entries and shows clear long-context gains.}
\resizebox{0.9\textwidth}{!}{
  
\begin{tabular}{c|c|ccc|ccc|ccc|ccc}
\toprule
      &  & \multicolumn{3}{c|}{16 views} & \multicolumn{3}{c|}{32 views} & \multicolumn{3}{c|}{64 views} & \multicolumn{3}{c}{128 views} \\
Datasets & Method & AUC30 & AUC5 & AUC3 & AUC30 & AUC5 & AUC3 & AUC30 & AUC5 & AUC3 & AUC30 & AUC5 & AUC3 \\
\midrule
\multirow{7}[2]{*}{RE10k}
& Cut3R  & 70.6 & 24.9 & 14.9 & 75.1 & 31.7 & 20.2 & 77.1 & 36.0 & 24.0 & 75.5 & 33.2 & 20.9 \\
& Fast3R & 69.3 & 22.3 & 13.0 & 76.4 & 31.5 & 19.7 & 80.6 & 38.6 & 25.2 & 79.1 & 37.8 & 24.1 \\
& Flare  & 75.2 & 29.1 & 17.9 & 81.0 & 39.2 & 26.2 & 84.9 & 47.6 & 33.7 & 85.0 & 49.6 & 35.9 \\
& VGGT   & 75.5 & 25.3 & 14.0 & 78.0 & 30.2 & 17.9 & 79.5 & 33.2 & 20.1 & 80.2 & 34.1 & 20.6 \\
& TTT3R  & 73.7 & 27.6 & 16.6 & 78.3 & 35.7 & 23.6 & 79.9 & 40.2 & 28.0 & 76.8 & 35.9 & 23.8 \\
& DA3    & \underline{87.5} & \underline{53.4} & \textbf{40.9} & \underline{90.3} & \underline{61.7} & \underline{49.8} & \underline{92.0} & \underline{68.3} & \underline{57.8} & \underline{92.9} & \underline{72.4} & \textbf{62.3} \\
& Ours   & \textbf{88.3} & \textbf{53.8} & \underline{40.0} & \textbf{91.0} & \textbf{62.9} & \textbf{50.3} & \textbf{92.7} & \textbf{69.8} & \textbf{58.2} & \textbf{93.7} & \textbf{72.5} & \underline{61.3} \\
\midrule
\multirow{7}[2]{*}{Co3Dv2}
& Cut3R  & 67.6 & 25.7 & 15.9 & 76.3 & 35.2 & 22.4 & 76.9 & 32.6 & 18.5 & 73.7 & 25.1 & 12.2 \\
& Fast3R & 68.7 & 26.6 & 16.2 & 78.8 & 39.2 & 25.6 & 81.5 & 43.2 & 28.1 & 80.2 & 40.4 & 25.3 \\
& Flare  & 68.2 & 24.5 & 14.9 & 74.2 & 33.3 & 21.8 & 72.1 & 33.4 & 22.5 & 70.2 & 34.2 & 23.8 \\
& VGGT   & 80.2 & 45.2 & 33.9 & 86.8 & 58.6 & 47.2 & 90.0 & 67.2 & 57.0 & 91.3 & 71.4 & 62.1 \\
& TTT3R  & 70.3 & 28.4 & 17.8 & 78.8 & 38.9 & 25.6 & 79.3 & 37.8 & 23.2 & 78.1 & 32.8 & 18.3 \\
& DA3    & \textbf{85.2} & \textbf{57.4} & \textbf{47.2} & \textbf{90.5} & \textbf{69.3} & \textbf{60.0} & \textbf{92.3} & \textbf{73.5} & \underline{64.0} & \textbf{93.0} & \underline{75.3} & \underline{65.7} \\
& Ours   & \underline{82.2} & \underline{50.1} & \underline{39.2} & \underline{88.9} & \underline{65.1} & \underline{55.4} & \underline{91.8} & \underline{72.8} & \textbf{64.3} & \underline{92.9} & \textbf{76.6} & \textbf{68.8} \\
\bottomrule
\end{tabular}
}
\label{tab:pose}
\end{table*}%

%% file: tables/dl3dv-pose.tex
\begin{table*}[t]
\centering
\caption{Pose Estimation results on DL3DV-10K across 16–256 input views. The \textbf{best} is marked in bold, and the \underline{second best} is marked with underline. Our method demonstrates particularly strong gains on long sequences.}
\label{tab:dl3dvpose}
\resizebox{1.0\textwidth}{!}{

\begin{tabular}{c|ccc|ccc|ccc|ccc|ccc}
\toprule
      & \multicolumn{3}{c|}{16 views} & \multicolumn{3}{c|}{32 views} & \multicolumn{3}{c|}{64 views} & \multicolumn{3}{c|}{128 views} & \multicolumn{3}{c}{256 views} \\
Method & AUC30 & AUC5 & AUC3 & AUC30 & AUC5 & AUC3 & AUC30 & AUC5 & AUC3 & AUC30 & AUC5 & AUC3 & AUC30 & AUC5 & AUC3 \\
\midrule
Cut3R  & 88.2 & 60.5 & 48.8 & 89.0 & 66.2 & 55.5 & 88.2 & 64.1 & 52.2 & 82.1 & 48.2 & 34.1 & 73.0 & 31.4 & 19.2 \\
Fast3R & 75.2 & 34.0 & 22.7 & 79.5 & 41.9 & 28.8 & 75.9 & 39.6 & 26.9 & 64.5 & 24.7 & 14.8 & 52.4 & 15.3 & 8.0 \\
Flare  & 90.4 & 69.9 & 60.6 & 93.2 & 80.0 & 73.6 & 93.8 & 82.5 & 76.2 & 92.8 & 80.8 & 74.0 & 92.8 & 79.1 & 71.1 \\
VGGT   & 94.7 & 81.9 & 74.7 & 96.3 & 88.8 & 84.1 & {96.3} & \underline{89.4} & 85.5 & 95.2 & 88.2 & 84.1 & 95.2 & 87.9 & 83.6 \\
TTT3R  & 89.3 & 63.3 & 51.4 & 89.9 & 68.8 & 58.4 & 89.3 & 68.8 & 58.7 & 85.3 & 58.8 & 46.5 & 83.3 & 50.7 & 36.5 \\
DA3    & \underline{96.2} & \underline{88.7} & \underline{84.0} & \underline{96.6} & \underline{90.7} & \underline{86.5} & 96.0 & 90.0 & \underline{86.1} & \underline{95.5} & \underline{89.1} & \underline{84.5} & \underline{95.3} & \underline{88.5} & \underline{83.8} \\
Ours   & \textbf{96.9} & \textbf{91.6} & \textbf{88.4} & \textbf{97.3} & \textbf{94.0} & \textbf{92.3} & \textbf{96.3} & \textbf{93.5} & \textbf{92.0} & \textbf{95.6} & \textbf{92.0} & \textbf{90.0} & \textbf{95.7} & \textbf{92.1} & \textbf{90.1} \\
\bottomrule
\end{tabular}
}

\end{table*}

%% file: tables/dl3dv-nvs.tex
\begin{table*}[t!]
\caption{Novel view synthesis on DL3DV-10K (PSNR↑ / SSIM↑ / LPIPS↓) across 16–256 input views. The \textbf{best} is marked in bold. DepthSplat requires GT poses while others are pose-free. DepthSplat fails to run on more than 64 views.}
\label{tab:viewsyn}
\centering
\resizebox{0.9\textwidth}{!}{

\begin{tabular}{c|ccc|ccc|ccc|ccc}
\toprule
      & \multicolumn{3}{c|}{16 views} & \multicolumn{3}{c|}{64 views} & \multicolumn{3}{c|}{128 views} & \multicolumn{3}{c}{256 views} \\
Method & PSNR & SSIM & LPIPS & PSNR & SSIM & LPIPS & PSNR & SSIM & LPIPS & PSNR & SSIM & LPIPS \\
\midrule
DepthSplat & \underline{24.80} & \textbf{0.849} & \textbf{0.134} & 18.39 & 0.617 & 0.363 & -- & -- & -- & -- & -- & -- \\
AnySplat   & 23.15 & 0.757 & 0.158 & \underline{20.52} & \underline{0.643} & \underline{0.248} & \underline{19.45} & \underline{0.593} & \underline{0.294} & \underline{18.55} & \underline{0.543} & \underline{0.339} \\
Ours       & \textbf{25.77} & \underline{0.811} & \underline{0.152} & \textbf{24.33} & \textbf{0.752} & \textbf{0.182} & \textbf{23.31} & \textbf{0.710} & \textbf{0.208} & \textbf{22.09} & \textbf{0.654} & \textbf{0.250} \\
\bottomrule
\end{tabular}
}
\end{table*}

%% file: tables/re10k.tex
\begin{table*}[t]
\centering
\caption{NVS on RE10K dataset. We do not use test time pose alignment for a fair comparison. We highlight the \textbf{best} and the \underline{second best}. "*" indicates methods require known intrinsics.}
\label{tab:re10k}
\resizebox{0.9\textwidth}{!}{
\setlength{\tabcolsep}{1.0pt} %

\begin{tabular}{lcccc}
\hline
                       & \multicolumn{4}{c}{Overlap Settings}                                                                                                                              \\ \cline{2-5} 
Method                 & Small                                  & Medium                                 & Large                                  & Average                                \\
                       & PSNR$\uparrow$ SSIM$\uparrow$ LPIPS$\downarrow$                        & PSNR$\uparrow$ SSIM$\uparrow$ LPIPS$\downarrow$                      & PSNR$\uparrow$ SSIM$\uparrow$ LPIPS$\downarrow$                       & PSNR$\uparrow$ SSIM$\uparrow$ LPIPS$\downarrow$                        \\ \hline
\textbf{Pose-Required} &                                        &                                        &                                        &                                        \\
pixelNeRF              & 18.417 0.601 0.526                     & 19.930 0.632 0.458                     & 20.869 0.639 0.485                     & 19.824 0.626 0.485                     \\
AttnRend               & 19.151 0.663 0.368                     & 22.532 0.763 0.186                     & 25.897 0.845 0.269                     & 22.664 0.762 0.269                     \\
pixelSplat             & 20.263 0.717 0.266                     & 23.711 0.809 0.181                     & 27.151 \underline{0.879} 0.122                     & 23.848 0.806 0.185                     \\
MVSplat                & 20.353 0.724 0.250                     & \underline{23.778} \underline{0.812} \underline{0.173}                     & \textbf{27.408} \textbf{0.884} \underline{0.116}                     & 23.977 0.811 0.176                     \\ \hline
\textbf{Pose-Free}     &                                        &                                        &                                        &                                        \\
DUSt3R                 & 14.101 0.432 0.468                     & 15.419 0.451 0.402                     & 16.427 0.453 0.432                     & 15.382 0.447 0.432                     \\
MASt3R                 & 13.534 0.407 0.494                     & 14.945 0.436 0.418                     & 16.028 0.444 0.452                     & 14.907 0.431 0.452                     \\
Splatt3R               & 14.352 0.475 0.472                     & 15.529 0.502 0.425                     & 15.817 0.483 0.421                     & 15.318 0.490 0.436                     \\
CoPoNeRF               & 17.393 0.585 0.462                     & 18.813 0.616 0.392                     & 20.464 0.652 0.318                     & 18.938 0.619 0.388                     \\
SelfSplat              & 17.506 0.550 0.461                     & 19.357 0.704 0.378                     & 20.868 0.672 0.256                     & 19.33 0.656 0.363                      \\
NoPoSplat*              & 21.814 0.765 0.220                     & 23.044 0.787 0.178                     & 25.408 0.844 0.126                     & 23.424 0.798 0.173                     \\
CocaSplat           & 21.312 0.741 0.234                     & 22.891 0.773 0.189                     & 25.312 0.829 0.138                     & 23.200 0.781 0.185                     \\
CocaSplat*              & \underline{22.843} \textbf{0.781} \textbf{0.192}                     & 23.394 0.804 0.175                     & 26.270 0.872 \textbf{0.110}                    & \underline{24.093} \textbf{0.818} \textbf{0.160}                     \\
Flare              & 20.779 0.695 0.247                     & 22.412 0.743 0.193                   & 23.791 0.774 0.151                    & 22.404 0.740 0.194                     \\
Ours - F.T.            & \textbf{22.951} \underline{0.768} \underline{0.203}                     & \textbf{25.047} \textbf{0.814} \textbf{0.163}                    & \underline{27.193} 0.856 0.129                     & \textbf{25.146} \underline{0.815} \underline{0.164}                     \\ \hline
\textbf{Zero-Shot}     & \multicolumn{1}{l}{}                   & \multicolumn{1}{l}{}                   & \multicolumn{1}{l}{}                   & \multicolumn{1}{l}{}                   \\
AnySplat               & \multicolumn{1}{l}{15.171 0.587 0.412}                   & \multicolumn{1}{l}{17.352 0.622 0.344}                   & \multicolumn{1}{l}{19.828 0.663 0.277}                   & \multicolumn{1}{l}{17.526 0.625 0.341} \\
Ours                   & \multicolumn{1}{l}{\textbf{20.664 0.693 0.262}} & \multicolumn{1}{l}{\textbf{22.537 0.746 0.211}} & \multicolumn{1}{l}{\textbf{24.026 0.773 0.180}} & \multicolumn{1}{l}{\textbf{22.501} \textbf{0.741} \textbf{0.214}} \\ \hline
\end{tabular}
}

\end{table*}

%% file: sections/5-conclusion.tex
\section{Conclusion and Limitations}
We presented LVSPM, a unified framework for pose-free view synthesis from long unposed image sequences. LVSPM jointly estimates world-aligned camera poses and synthesizes novel views, trained with RGB images and camera-pose supervision, avoiding dense 3D geometric supervision while matching or outperforming methods that rely on it. Enabled by an efficient test-time training mechanism, LVSPM scales to sequences of 200+ views and maintains strong rendering quality under the practical equi-temporal evaluation protocol, where additional input views correspond to larger scene coverage. Across challenging benchmarks such as DL3DV-10K, this leads to graceful scaling as scene complexity increases, while existing baselines degrade more rapidly. LVSPM represents a significant step toward practical, scalable 3D scene understanding from unposed image collections, bringing us closer to systems that can operate on real-world data without expensive preprocessing or annotation requirements.

Despite these advances, several limitations remain. The multi-stage training pipeline introduces additional complexity. Performance may degrade under extreme lighting changes or in weakly textured scenes, where reliable geometric reasoning is difficult. In addition, although LVSPM achieves competitive pose estimation while jointly supporting view synthesis, specialized geometry-focused models may still perform better in certain edge cases. Future work includes exploring fully self-supervised training to further reduce supervision requirements and investigating how far scalable 3D scene understanding can be pushed with minimal geometric priors.